\documentclass[10pt,twocolumn,letterpaper]{article}

\usepackage{iccv}
\usepackage{times}
\usepackage{epsfig}
\usepackage{graphicx}
\usepackage{amsmath}
\usepackage{amssymb}
\usepackage{url}
\usepackage[pagebackref=true,breaklinks=true,letterpaper=true,colorlinks,bookmarks=false]{hyperref}

\iccvfinalcopy

\ificcvfinal\fi

\newcommand{\bftab}{\fontseries{b}\selectfont}

\begin{document}

\title{AdaptiveShape: Solving Shape Variability for 3D Object Detection with Geometry Aware Anchor Distributions}

\author{Benjamin Sick\\
{\tt\small benjamin.sick@zf.com}
\and
Michael Walter\\
{\tt\small michael.walter5@zf.com}\\
\\
{ZF Friedrichshafen AG}
\and
Jochen Abhau\\
{\tt\small jochen.abhau@zf.com}
}

\maketitle
\ificcvfinal\thispagestyle{empty}\fi

%%%%%%%%% ABSTRACT
\begin{abstract}
   3D object detection with point clouds and images plays an important role in perception tasks such as autonomous driving. Current methods show great performance on detection and pose estimation of standard-shaped vehicles but lack behind on more complex shapes as \eg semi-trailer truck combinations. 
   Determining the shape and motion of those special vehicles accurately is crucial in yard operation and maneuvering and industrial automation applications.
   This work introduces several new methods to improve and measure the performance for such classes.
   State-of-the-art methods are based on predefined anchor grids or heatmaps for ground truth targets. However, the underlying representations do not take the shape of different sized objects into account. 
   Our main contribution, AdaptiveShape, uses shape aware anchor distributions and heatmaps to improve the detection capabilities. For large vehicles 
   we achieve +10.9\% AP in comparison to current shape agnostic methods.
   Furthermore we introduce a new fast LiDAR-camera fusion. It is based on 2D bounding box camera detections which are available in many processing pipelines.
   This fusion method does not rely on perfectly calibrated or temporally synchronized systems and is therefore applicable to a broad range of robotic applications.
   We extend a standard point pillar network to account for temporal data and improve learning of complex object movements.
   In addition we extended a ground truth augmentation to use grouped object pairs to further improve truck AP by +2.2\% compared to conventional augmentation. 
\end{abstract}

%%%%%%%%% BODY TEXT
%-------------------------------------------------------------------------
\section{Introduction}

\begin{figure*}[t]
	\begin{center}
		\includegraphics[width=0.8\linewidth]{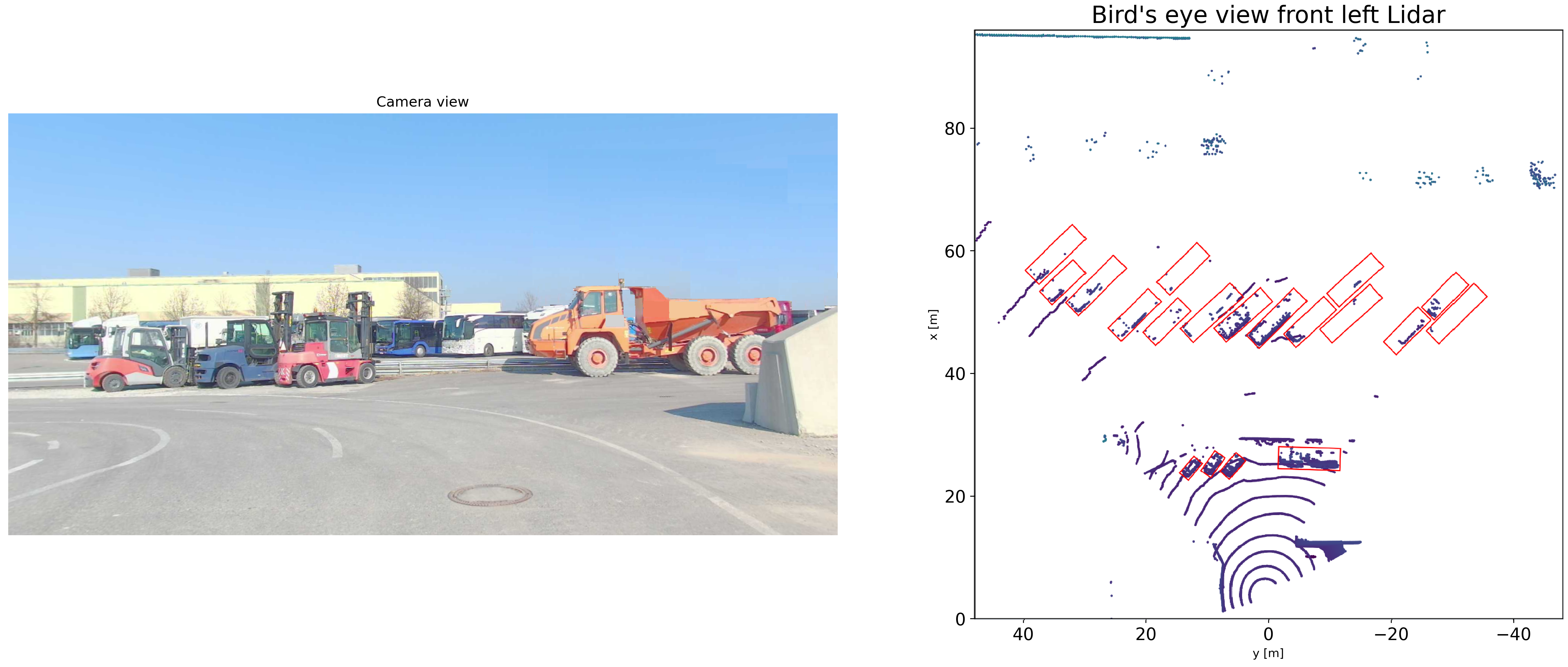}
	\end{center}
	\caption{
		Left: Premises with special non-standard vehicles like forklift and special truck in the foreground. Right: Bird's eye view of LiDAR pointcloud with ground truth boxes. Buses in the background with varying occlusions which make it impossible to determine the length from the point cloud.
	}
	\label{fig:other_vehicles}
\end{figure*}

Autonomous driving
requires the accurate detection of road users in the vicinity of the ego vehicle. 
Those objects can be divided in pedestrian, two-wheeler and vehicle with very different appearances. 
For example, the vehicle category contains large variations between a small car and a semi-trailer truck combination which needs to be accommodated for. Especially in industrial applications for yard operation and automation, the accurate detection of large vehicles is mandatory. 
Container terminals or company premises are examples where 
special vehicles with quite different shapes like semi-trailer truck combinations, tractors and forklifts
are maneuvering which results in lots of different rotations and constellations, see Figure \ref{fig:other_vehicles}. It was recently shown in \cite{RareExamples} how important it is to include rare objects.
3D object detection is usually done with 7 degrees of freedom bounding boxes containing the 3D coordinates, width, length, height and rotation around the Z-axis of the objects. The three sensor modalities camera, LiDAR and radar are mainly used for perception. Each sensor has its strengths and weaknesses. 
Camera has the highest resolution and can capture textures. However it is a passive sensor with drawbacks in bad lightning. It can not estimate the distance of objects reliably with high precision unless a precisely calibrated wide base stereo system is used.
Radar has a very high range and can measure the radial velocity of objects with very high precision. However it has a low angular resolution and suffers from sidelobes which can mask weak targets.
LiDAR has, compared to radar, a much higher angular resolution. 
Latest models have a resolution of around 0.1 degree and can detect even small objects.

3D object detection benchmarks are dominated by LiDAR only methods which are sometimes surpassed by LiDAR-camera fusion models. However the higher the LiDAR resolution and the lower the number of classes which need to be classified, the smaller the benefit of the camera becomes (compare \eg Waymo Open \cite{Waymo} vs. nuScenes \cite{nuScenes}). Camera can mainly help to classify objects into precise categories as \eg motor-cycles vs. bicycles, detect small targets in greater distances or differentiate pedestrians from other similar shapes as \eg poles or trash cans. 

Early neural network implementations were based on Pointnets \cite{PointNet, PointNetPlusPlus} and were soon augmented with camera information by \eg Frustum-PointNet \cite{FrustumPointNet} which crops points inside the camera frustum of 2D bounding boxes. A 3D detection based on segmentation of the point cloud into foreground and background can also be used \cite{PointRCNN}.
Since Voxelnet \cite{VoxelNet} as a generic one-stage model, most methods use a voxelized point cloud representation \cite{Polarnet,Cylindrical}. 
The projected range image for object detection is used by \cite{RangeSparseNet,RangeDet}. Hybrid feature methods including the voxel network are \cite{PVRCNN,HybridVoxel,HVNet}.
Due to the need of 3D convolutions the Voxelnet approach was slow at the beginning until SecondNet \cite{SecondNet} introduced sparse 3D convolutions from \cite{Sparse3dConv}. SecondNet also added ground truth database sampling and focal loss \cite{FocalLoss} to improve performance.
Based on this, Point Pillars \cite{PointPillars} introduced a pillar based approach which operates in bird's eye view (BEV) and was able to operate faster than real time.
The Point Pillars network was extended in \cite{ClassBalanced} to include multiple detection heads and a class balanced sampling for different classes.

3D object detection methods are often divided into anchor-based \cite{PointPillars,SecondNet,PointRCNN} and anchor-free \cite{AFDet,PillarBased,CenterNet} methods. In our view anchor-free is however a misnomer. "Anchor-free" methods use anchors which do not rely on the IoU with predefined anchor bounding boxes.
\cite{CenterPoint} introduced the heatmap based detection inspired from image processing \cite{CornerNet} and \cite{CenterNet}.
Fusion of camera images can be done inside one network \eg using a unified BEV representation \cite{BEVFusion} or as a point cloud annotation \eg \cite{PointPainting}.
Most recently networks based on the attention mechanism in transformer networks \cite{Attention} are being used to extract more complex scene relations \cite{CenterFormer}.
Class imbalances can be handled by resampling \cite{ClassBalanced}, adopted weighting \cite{EffectiveSamples}) or
loss function adaptation \cite{FocalLoss, LongTailVisual}. Knowledge transfer \cite{FeatureSpace, LearningTail} or expert learning and distillation \cite{LearningExperts, Distillation} are further methods.

Our work is loosely based on the \cite{ClassBalanced} paper for the new anchor based AdaptiveShape implementation to make use of the fast inference speed of the PointPillars architecture. For the heatmap based method we based our work on \cite{CenterFormer}. 

\paragraph{Contributions}
\begin{itemize}
	\item Our main contribution is a new shape aware anchor and heatmap generation method, AdaptiveShape, which improves detection performance for datasets with a large variance in object geometries.
	\item LiDAR camera fusion concept which is fast and robust against calibration errors or temporally unaligned sensors.
	\item Multi-frame temporal stacking concept which allows to extract movement patterns from objects.
	\item Adapted evaluation metric to improve analysis of underrepresented classes.
	\item Paired ground truth database sampling for point cloud augmentation which improves the augmentation for object pairs.
\end{itemize}

%-------------------------------------------------------------------------
\section{Anchors and heatmap}

\begin{figure*}[t]    
	\begin{center}        
		\includegraphics[width=0.7\linewidth]{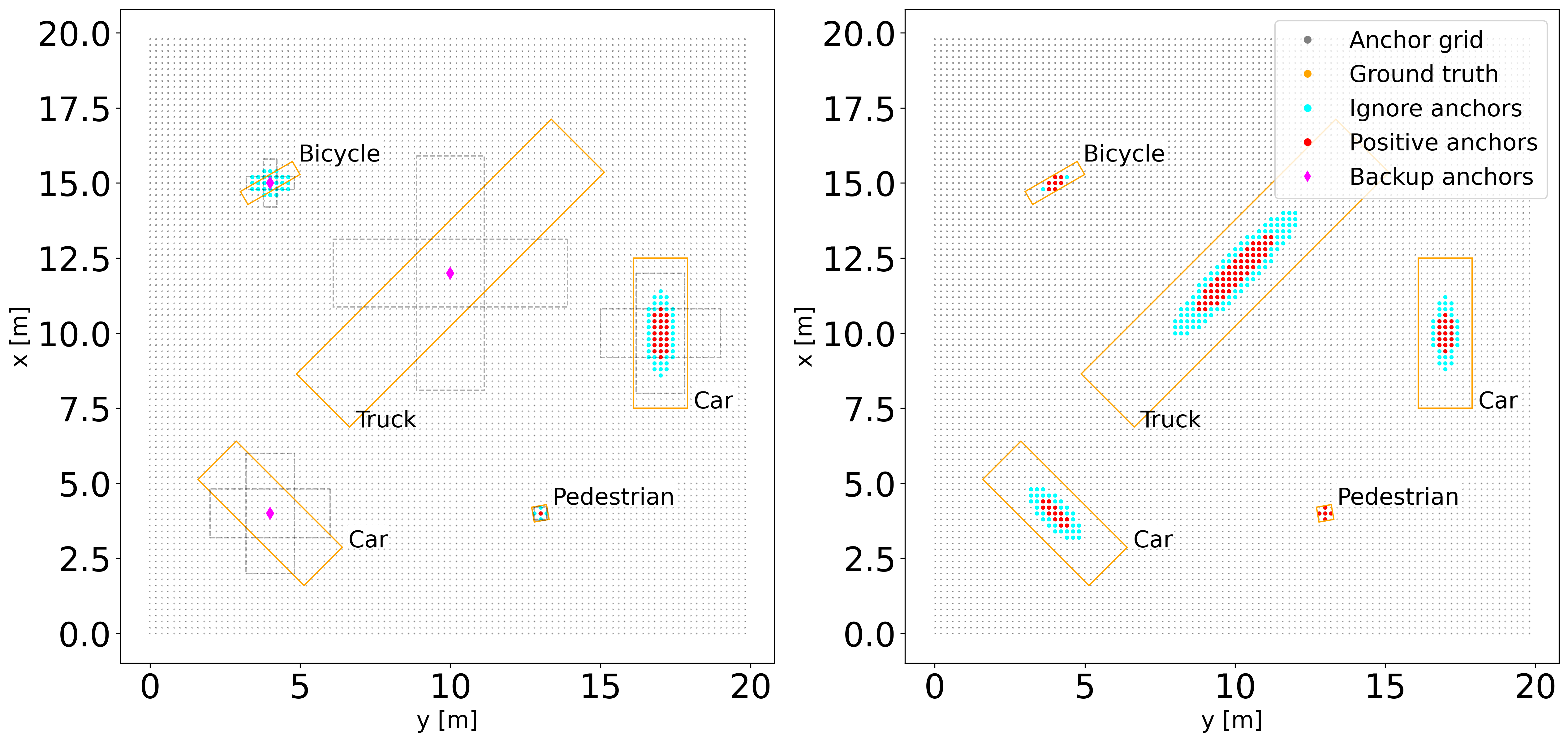}    
	\end{center}    
\caption{        
	Left: Original PointPillars anchor distribution. Anchors are created for 0 and 90 degree for each class and matched based on an axis aligned IoU with ground truth. The anchor distribution is highly dependent on the orientation and class dependent anchor box size.
	Right: Our new AdaptiveShape implementation which distributes anchors in an rotated ellipse around the center. The anchor distribution is independent of object category standard sizes and orientations. Large objects which are usually underrepresented will get more anchors.    
}    
\label{fig:anchors}
\end{figure*}

Anchor grids are a well-established method for BEV bounding box regression. Benjin Zhu et al. \cite{ClassBalanced} use two anchor grids for each object category. Each of those grid cells comprises anchor boxes with two orientations, 0 and 90 degrees and a mean category dimension.
Foreground/background and ignore anchors are chosen based on two thresholds. An anchor is classified as foreground anchor if the intersection over union (IoU) between the ground truth and anchor box is above a first threshold. An anchor is classified as a background anchor if the IoU is below a second threshold. If the IoU is in between the first and second threshold an anchor is considered an ignore anchor, which won’t be accounted for in the loss.
This leads to problems as can be seen in Figure \ref{fig:anchors}. Objects with orientations close to 45 degree or with unusual object dimensions get no positive anchors assigned. If no anchor is assigned to an object, a fallback anchor in the center of the object is used. However, this fallback leads to a single binary classification without ignore area. Moreover it neither accounts for the object shape nor the object orientation.
On the other hand objects with standard dimensions and orientations (as normally seen from a vehicle perspective) similar to the anchors get many positive anchors assigned as can be seen for the car with 0 degree orientation in Figure \ref{fig:anchors} on the left. Therefore a large imbalance is created which favors those standard objects with standard orientations. Usually, those objects are cars driving in the front or parked on the side. 

Trucks and trailers on the other hand have largely varying shapes, especially lengths. All orientations, not only preceding or crossing ones, are seen in yard automation, as depicted in Figure \ref{fig:other_vehicles}. The result is that they get most of the time only the backup anchor assigned. 

Some public 3D object detection challenges contain a disproportional high amount of the favored standard objects. Metrics usually do not explicitly account for different orientations therefore the typical high performing object detection algorithms are overfitted on those standard objects and perform worse on special vehicles with less frequent orientations. 

One way of mitigation is the approach presented in \cite{PillarBased} or CenterPoint \cite{CenterNet}\cite{CenterPoint}. As described in \cite{CenterNet}: "A center point can be seen as a single shape-agnostic anchor". In fact, CenterPoint is an extension to the backup anchor strategy from the original anchor approach that takes the local neighborhood into account and replaces the discrete foreground/background/ignore classes with a distance-based weighting. 
Anchors will be weighted with a continuous loss according to the distance from the center. Nevertheless, the original CenterNet and 3D CenterPoint are not really shape aware since they do not use different standard deviations for length and width and use only an uncorrelated Gauss distribution that does not account for the object orientation. 
Different lengths and widths are mentioned in the papers but the source code at the respective repositories does not create a correlated distribution \cite{CenterPointGithub}. 

We show that it is an advantage to have multiple anchors representing the shape and orientation of a single object.
Having multiple anchors, the network must determine the visible corners of an object by estimating the offsets to a virtual center and virtual length, inferred from similar objects in the training set, for multiple anchor positions. Moreover, having a Gaussian anchor distribution will generate an intermediate representation that more closely resembles the shape of an object. 

We can reduce the anchor size to a norm size of \eg 1x1m and calculate an intersection over the anchor size instead of an intersection over union. This leads to multiple positive anchor positions within an object and scales with its size. 
However, we encountered interpolation artifacts at the object boundary of two overlapping vehicles, like a truck semi-trailer combination. 
Therefore it would require a median based post processing for the results of the regression head or an additional distance restriction from the center of an object where anchors will be considered. 

Here we present two shape aware methods, named AdaptiveShape: The first one is an extension of a classical anchor grid using foreground and ignore anchors. The second one is an extension of CenterNet where we replaced the uncorrelated Gaussian with a fully correlated shape/orientation dependent Gauss distribution.

\subsection{Shape aware grid representation}
 
Our new grid based approach embeds two ellipses based on two fully correlated Gauss distributions encoding the shape and orientation of an object, one for the positive and one for the ignore anchors, which are ignored in the loss function.
Therefore anchors are distributed in an rotated ellipse around the center. The length and width of the ellipse corresponds to the box ratio and the rotation of the ellipse is the same as the box orientation. 
Anchors are thereby restricted not to be too far away from the center. This is especially important for overlapping objects to avoid \eg interpolations at the transitions between trucks and semi-trailers. 
A pleasant side effect of this approach is, that the network is trained on more positive anchors for large objects which are usually underrepresented. This helps to counteract this imbalance and is preferred to a stronger weighting of those classes in the loss. It forces the network to learn offsets for object edges for different anchor positions. 

\subsection{Fully correlated Gaussian heatmap} 

The original Gaussian heatmap implementation in \cite{CenterPoint} is based on an uncorrelated Gauss distribution which neither takes a different ratio between lengths and widths nor the object orientation into account.
Our approach improves this by using a fully correlated Gauss distribution with individual standard deviations based on the lengths and widths of objects. 
The Gaussian is rotated according to the orientation of the object. Therefore we achieve a similar effect as in the new anchor assignment, where elongated and/or rotated objects have an elongated and/or rotated Gaussian distribution. 

\begin{figure*}[t]
	\begin{center}        
		\includegraphics[width=0.7\linewidth]{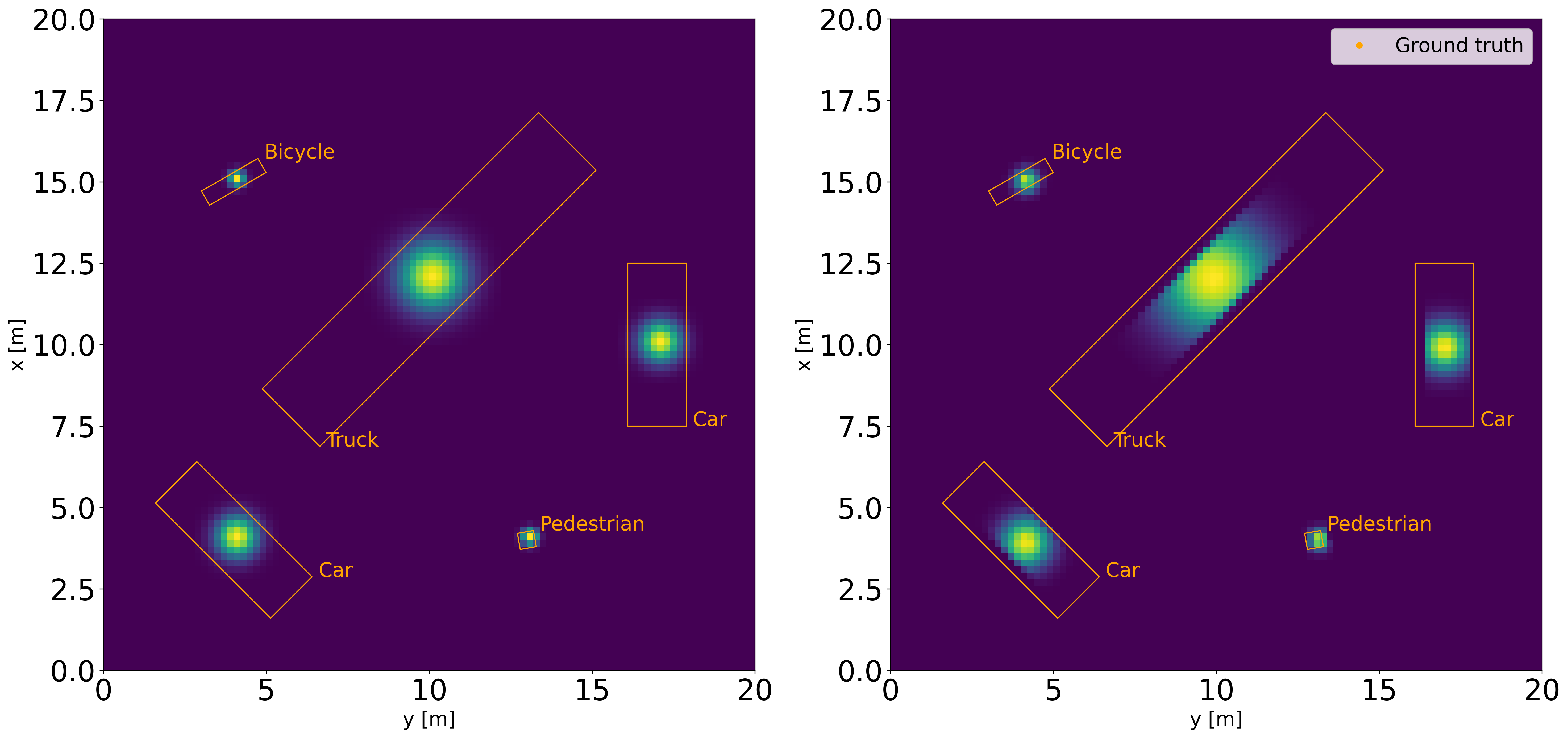}    
	\end{center}    
\caption{        
	Left: Original uncorrelated CenterPoint Gauss distributions which do not take the orientation or the length/width ratio into account. All distributions have the same standard deviation in both directions. Right: Our new AdaptiveShape implementation which takes orientation and length/width ratio of objects into account. For long objects like the truck in the center we can see how the distribution is elongated along the length of the truck. 
}    
\label{fig:gaussians}
\end{figure*}

%-------------------------------------------------------------------------
\section{LiDAR camera fusion}

\begin{figure*}[t]
	\begin{center}
		\includegraphics[width=0.7\linewidth]{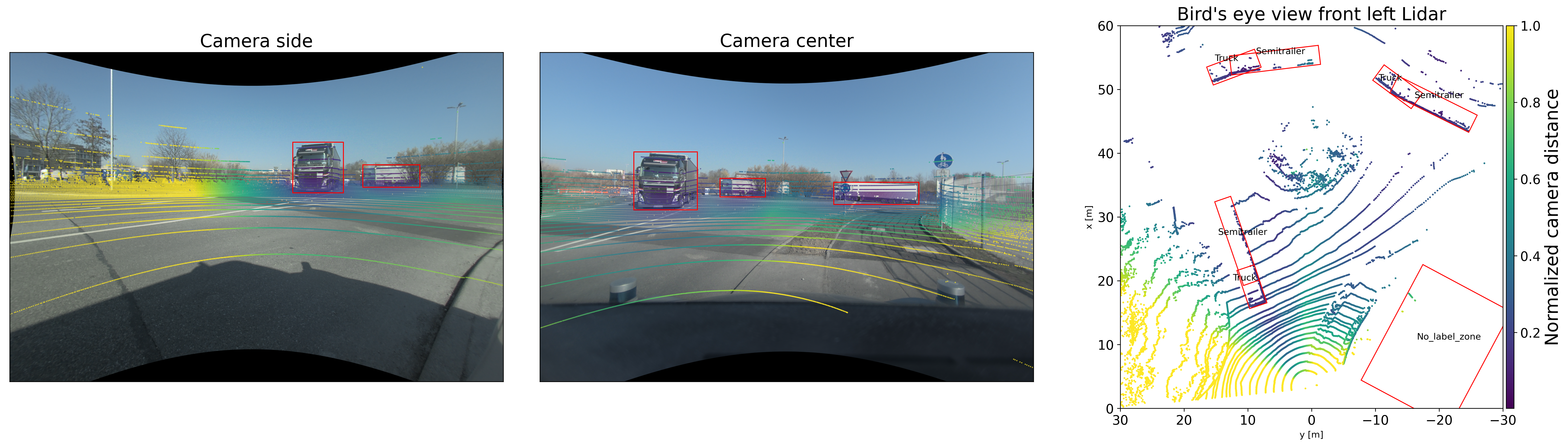}
	\end{center}
	\caption{
		Camera LiDAR PointDistanceFusion based on the distance of the backprojected LiDAR points to nearest camera object in camera projection. The distance is added to the original point cloud as an extra column and allows the network to improve detection performance and makes it more robust against calibration and synchronization errors.
	}
	\label{fig:cameradist}
\end{figure*}

At the same time when Sourabh Vora et al published their PointPainting \cite{PointPainting}, we came up with an identical approach based on the segmented image but switched soon to a bounding box classifier based image augmentation, in the following referred to as PointDistanceFusion.
PointPainting projects the point cloud onto a segmented image and individual points are classified based on their object category as foreground or background points. The main problem with this approach however is, that the complete system needs to be perfectly synchronized and calibrated. Otherwise, the overlaid point cloud will not fit to the camera objects and many background points will be classified as being part of a foreground object and vice versa.
Our method is based on the distance between the back projected points and the center of the classified 2D bounding boxes, see Figure \ref{fig:cameradist}.
Using the vehicle ego motion, we transform the points to the camera shutter timestamp and project the point cloud onto the image. In the image coordinate system, we calculate the distance of each point to its nearest camera bounding box for the different object categories. The distance is subsequently normalized by the width/height of the bounding box (thus encoding the distance in 3D), which is then used as an extra column in the point cloud. 
Optionally the camera based object depth can be appended.
Using a distance measure makes us more robust to calibration and synchronization errors, especially when it comes to moving objects with unknown speed and direction of movement. In addition, while PointPainting relies on an object segmentation network, PointDistanceFusion utilizes a bounding box-based classification network as found in most autonomous driving perception stacks. 

%-------------------------------------------------------------------------
\section{Fusion of multiple temporal frames}
\label{sec:temporal}
To improve the confidence of an object and derive additional measurements like the velocity or yaw rate, we consider the last 3-frames together. The simplest way to account for multiple frames is to augment each feature vector within an ego motion compensated frame with the time offset to the current frame \eg \cite{nuScenes}.

\begin{figure}[t]
	\begin{center}
		\includegraphics[width=0.8\linewidth]{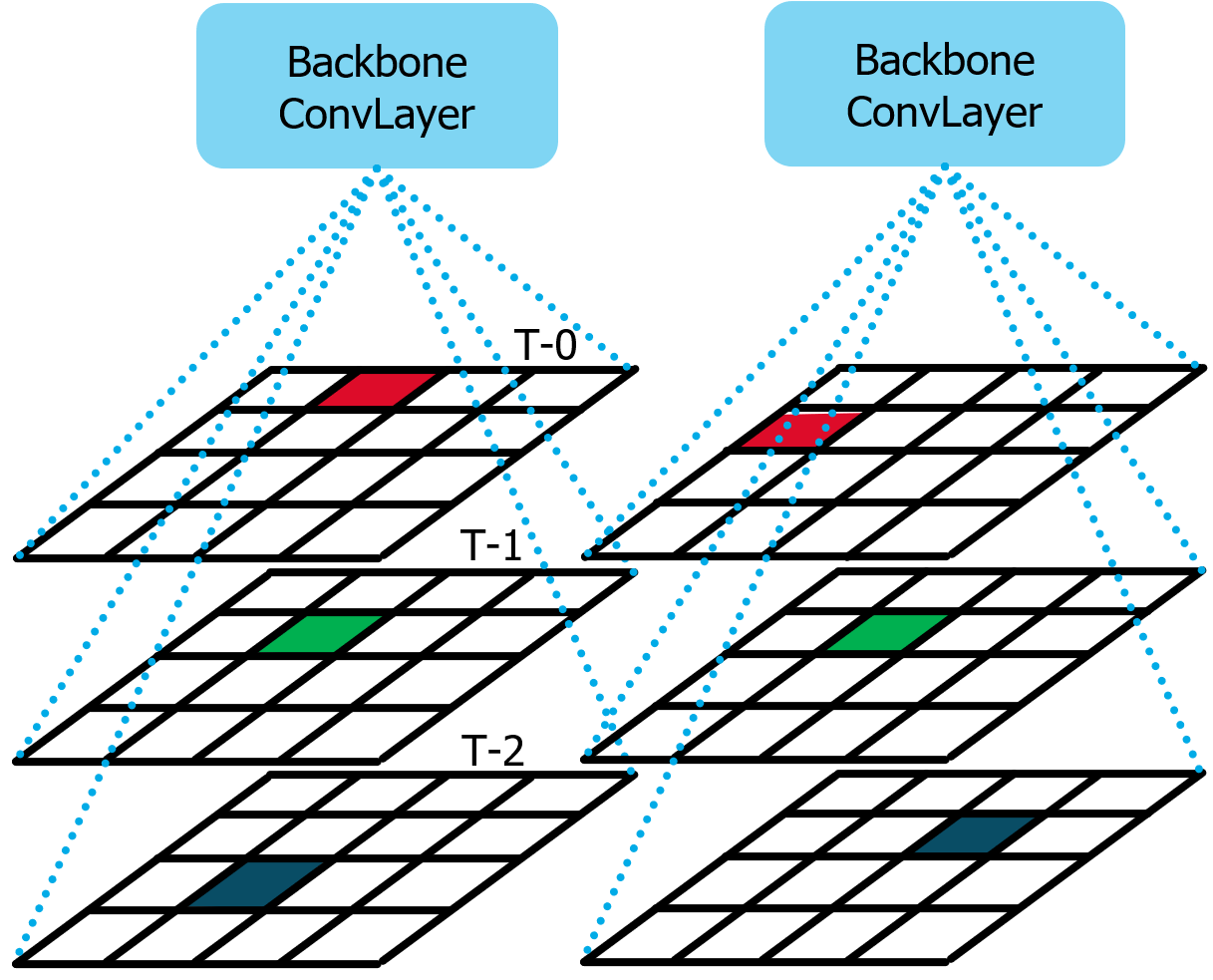}
	\end{center}
	\caption{
		We can derive temporal information like the velocity or yaw rate of a vehicle once the ego motion compensated features are stacked on top of each other using a simple convolutional filter. On the left we can see the temporal pattern of an object moving from bottom to top. On the right we can see an object moving from right to left.
	}
	\label{fig:temporalB}
\end{figure}

However, since our network is based on a point pillar architecture it is not clear how a simple augmentation will be propagated through it. 
Therefore, we decided to modify the architecture in such a way that the output of the ego motion compensated features can be stacked on top of each other as separate layers once the output of the point pillar net is mapped onto a grid representation. This allows the network to learn more complex patterns using simple 2D convolutions, spanning over all temporal layers, as seen in Figure \ref{fig:temporalB}. 

Having a point pillar architecture, we extend the shape of the input [BatchSize, NrOfPillars, NrOfPointsPerPillar, NrOfFeatures0] to account for the last N frames [BatchSize, N , NrOfPillars, NrOfPointsPerPillar, NrOfFeatures0]. This input is reshaped to [BatchSize * N, NrOfPillars, NrOfPoints, NrOfFeatures0] and can be directly used as an input for a standard point pillar network. A pleasant side effect is, that the number of training examples seen by the network is increased by the number of time steps to be considered. Once the output of the point pillar network is mapped onto a bird's eye view grid representation [BatchSize * N, GirdSizeX, GridSizeY, NrOfFeatures1] we reshape the output to a tensor having a size of [BatchSize, GirdSizeX, GridSizeY, N * NrOfFeatures1]. The ego motion compensated temporal features are stacked on top of each other which allows the network to extract temporal information using a simple convolutional network.
To correctly sample objects, the ground truth of the previous frames must be taken into account. If frames are not continuously labeled, interpolation can be applied, or the training set can be augmented using simulated data.

%-------------------------------------------------------------------------
\section{Ground truth analysis and adapted metrics}

In our opinion an often overlooked topic is the in depth analysis of the ground truth dataset and the fine tuning of the used metrics. Having investigated point clusters of long vehicles and trailers, we see that there is often no information regarding the length of the object because large objects are often partly occluded. Similar looking trailers from the front/back might have largely varying lengths. Using the IoU or bounding box center for ground truth matching for extended objects with non-observable lengths is not optimal as there might be large differences. A better approach can be to check for the alignment of the front/rear face of objects. This information is also more important for collision avoidance.

In addition metrics are usually calculated using the average precision (AP) for different object categories. Those categories can be fine as seen in \eg nuScenes \cite{nuScenes} or coarse as seen in \eg Kitti \cite{Kitti} or Waymo Open \cite{Waymo} datasets. 
The choice of categories highly influences the results. This explains \eg why there seems to be no benefit of the camera fusion for Waymo Open while for nuScenes having more classes and lower LiDAR resolution we can see a clear benefit.

A finer class partitioning introduces artificial boundaries between objects of similar shapes. SUVs, pickup trucks and vans can belong to the car or truck class.
Therefore the network is forced to overfit on class decision boundaries which are not visible in the data. This can significantly worsen the performance in a class specific metric while the actual detection capability of a network can be superior. In real world situations the accurate class of an vehicle is rarely needed. Class divisions should only be introduced for those cases. 
Additionally most of those vehicles behave similar in traffic apart from larger (semi-)trucks with trailers which have different orientations. 

The problem with a coarse class partitioning is that the performance for underrepresented classes is not measurable anymore. If \eg all vehicles are in one common class, the performance of this metric is mainly influenced by the most common vehicle type which is often medium sized cars. 
Significantly worse performance of an underrepresented subclass will not be visible if the overrepresented class has improvements.

Some methods try to overcome this class imbalance by resampling the ground truth and repeating frames with more underrepresented objects \cite{ClassBalanced}. 
Another method is the ground truth database sampling which we investigate deeper in section \ref{sec:gt}.

\subsection{Adapted metric}
\label{sec:metric}

To overcome these problems we propose a new metric calculation. Our metric is split in four different classes: car, truck, pedestrian and two-wheelers. However we allow mis-classifications of cars as trucks and vice versa (the same for cyclist and pedestrian classes). As an example, for the truck metric we check that for each truck ground truth we have either a corresponding truck or car class. For false positives, we check for each truck detection if a corresponding truck or car ground truth is available. This new metric allows to see changes for underrepresented classes while not artificially separating similar classes.

Additionally we use so called no label zones. Those zones are added to the ground truth in the following ways: 
During manual labeling at non relevant areas in the environment, \eg parking spaces or automatically based on point cloud distribution, \eg every ground truth box with less then 5 LiDAR points is set to an ignore zone.
Instead of removing all points inside the no label zones and the corresponding ground truth, we ignore them in the loss calculation. This has the advantage that the network can still see patterns containing ignore zones and improve its larger scene understanding. An example would be an occluded car in between two other cars, so the network won't be punished for detecting it.

%-------------------------------------------------------------------------
\section{Improving ground truth database sampling}
\label{sec:gt}

A common method for training of 3D object detection is the ground truth database sampling. A database is created which contains the points within the 3D bounding box and class information for each ground truth object. See \eg Figure \ref{fig:towing} on the left for such a point cluster. During training, each frame is augmented by pasting a random selection of objects from the database into the current frame.
Recent works introduced the fading technique where ground truth sampling is stopped early when the model is near convergent \cite{PointAugmenting}. This shows that the augmented point clouds do not represent perfectly valid scenes but rather kick start the training process with lots of examples.
A further performance improvement can be achieved by a more realistic ground truth database sampling. We propose several methods. 

\subsection{Sampling pairs of connected objects}

\begin{figure*}[t]
	\begin{center}
		\includegraphics[width=0.8\linewidth]{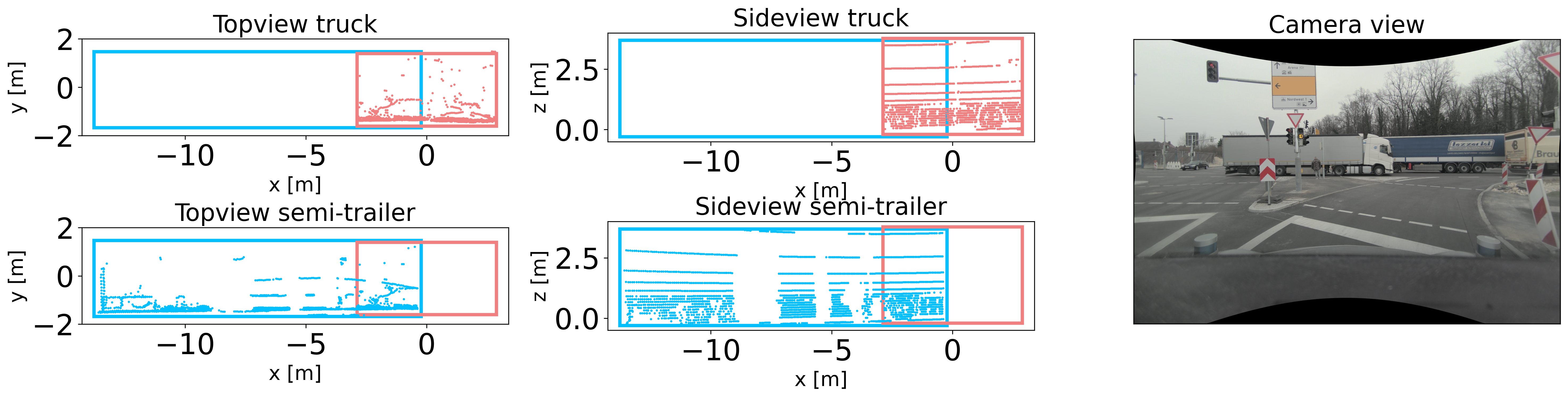}
	\end{center}
	\caption{
		Top- and side-view of a semi-trailer truck combination ground truth pair (truck in red, semi-trailer in blue). If we cut and paste only one of the two objects, we get deformed objects which do not resemble realistic vehicles. Therefore we store the pair relation at ground truth database creation and paste such objects together when augmenting frames during the training.
	}
	\label{fig:towing}
\end{figure*}

In our dataset we introduced an extra category named "towing vehicle" to accommodate the special nature of semi-trailer truck combinations, see Figure \ref{fig:towing} for an example. The movement of these vehicles is quite special in curves as the direction of the trailer will not point towards the driving direction. Only the truck part points in the actual driving direction and needs to be therefore detected separately to predict the future trajectory, see Figure \ref{fig:cameradist}. The difficulty for separated detection is that the semi-trailer and truck have a section where the point cloud overlaps in bird's eye view.
If we consider a semi-trailer truck with an attached semi-trailer but we cut-off the point cloud at the end of the truck it looks significantly different from a standalone semi-trailer truck, see Figure \ref{fig:towing}. During conventional ground truth database generation, the pair relation of those objects is not stored and therefore those objects are pasted separately during training, possibly in different frames. With a pairwise augmentation, PairAugment, we store this relation and check during training if objects from the database contain a corresponding pair relation. If this exists, we paste both objects together as they were also seen in the original data.

\subsection{Include correct points from temporal history}

To improve performance or when direction and velocity information is needed, multiple LiDAR frames can be stacked together for the network input, as discussed in section \ref{sec:temporal}. 
As we observed, most state of the art open source repositories do not fully account for the temporal history \cite{CenterPointGithub, CenterFormerGithub}. 
They only account for the history of the current bounding box but ignore the full history of the previous measurements. Therefore the sampled points from multiple temporal frames are artificially cut-off at the bounding box border while the bounding boxes of previous frames would extend further. 
To correctly sample objects, the ground truth of the previous frames must be taken into account. 
Interpolation can be used to get the full ground truth if continuous labels for all temporal frames are not available. 
We implemented a new ground truth sampling which takes previous frame bounding boxes into account.

%-------------------------------------------------------------------------
\section{Experiments}

As baseline, we re-implemented the ideas from \cite{ClassBalanced} and use an adapted PointPillars network with multiple detection heads and a slightly modified backbone.
For the camera LiDAR PointDistanceFusion we augmented the network input with two distance measures, one for vulnerable road users and one for vehicles.
Training is done with 24 epochs, batch size 8, Adam optimizer with weight decay \cite{AdamW} and one cycle learning rate scheduler \cite{OneCycle} and with a pillar size of 0.2 meters.
For the final detection list we use non-maximum supression. No ensembles or test time augmentation is used.
For an additional experiment we adapted the CenterFormer implementation from \cite{CenterFormerGithub} with a fully correlated Gaussian heatmap. The source code of the changes can be found in the supplementary material.

\subsection{Dataset}

Our dataset was recorded with different test vehicles between January 2019 and September 2022. It was mainly recorded in southern Germany and contains large variations in seasons, weather and daytime.
To be able to evaluate the performance of large vehicles it should be mentioned, that the dataset is focused on rather difficult truck semi-trailer combinations as well as special vehicles, \eg construction vehicles, see Figure \ref{fig:other_vehicles}. 
For this reason we recorded the scenes on company yards. Additionally, we recorded scenes at roundabouts next to company yards with lots of passing semi-trailer truck combinations (Figure \ref{fig:cameradist}).
The recording setup consists of three front facing fisheye cameras, two Velodyne VLP-32 LiDARS at the front left and right, one high resolution 4D imaging radar at the front center and one Oxford GPS and IMU.
The dataset contains around 90.000 frames with around 850.000 manually 3D labeled objects.
Labeling was done on small segments of four frames with 500ms separation. Segments were chosen based on a scene selection algorithm which was applied on the recorded data. 
The test set contains 11,415 frames and was chosen to have mostly different geographic locations but a similar class distribution to the training set. 

%-------------------------------------------------------------------------
\section{Results}

\subsection{Shape aware grid representation}

Table \ref{tab:results} shows our results in comparison to the baseline. Each new extension was evaluated with all previously ones applied.
Additionally, to show the effect of using only the fallback anchor, we run one experiment with PointDistanceFusion using only one single positive anchor in the center of the object.
To calculate average precision we use the nuScenes metric with the adaptations mentioned in \ref{sec:metric}. The maximum evaluated detection distance is 96 meter instead of the 40/50 meter used in nuScenes. We also report the mean average precision (mAP) over all categories.
To be more comparable to nuScenes we also calculated the average precision for a range of 40 meters for vulnerable road users and 50 meters for vehicles. 

\begin{table*}[t]
	\centering
	\begin{tabular}{lrrrrr}
		\hline
		Experiment                            &   Truck AP &   Pedestrian AP &   Cyclist AP &   Car AP &   mAP \\
		\hline
		\multicolumn{6}{c}{Evaluation up to distance 96m} \\
		Baseline IoU anchor                   &       29.3 &            78.4 &         62.8 &     88.9 &  64.9 \\
		+PointDistanceFusion                  &       29.7 &            80.6 &         63.7 &     89.5 &  65.9 \\
		+Center anchor                        &       26.2 &            80.9 &         56.1 &     89.4 &  63.2 \\
		+AdaptiveShape anchor                 &      \bftab 40.6 &           \bftab 81.5 &        \bftab 64.5 &    \bftab 89.9 & \bftab 69.1 \\
		\hline
		+Ground truth augmentation            &       43.7 &            80.8 &         63.4 &     88.7 &  69.2 \\
		+PairAugment                          &      \bftab 45.9 &            81   &         63.3 &     88.6 & \bftab 69.7 \\
		\hline
		\multicolumn{6}{c}{Evaluation up to distance of 50m (vehicles) and 40m (rest)} \\
		Our best model & 51.5 & 87.3 & 75.6 & 93.5 & 77.0 \\
		\hline
	\end{tabular}
	\caption{
		Experiment results. The largest improvements can be seen for the truck class using the new AdaptiveShape anchor generation. Camera fusion improves the results of all classes. The ground truth augmentation additionally improves the truck detection and is further improved when pairs are considered. This results in the overall best mAP. 
		To be more comparable to nuScenes we also evaluated our best model on our data with a evaluation distance up to 40/50m.
	}
	\label{tab:results}
\end{table*}

\paragraph{PointDistanceFusion}
The camera fusion improves the pedestrian detection by +2.2\% AP. The other classes are improved between +0.4\% and 1.0\% AP. This shows how the camera can be used to improve the detection of small objects which are only covered by few LiDAR points, especially pedestrians by adding a focus of attention.

\paragraph{Center anchor}
As expected using only one single anchor in the center of the object leads to a poor results for larger objects. All classes decrease between -0.1\% and -7.5\% AP, while only the pedestrian detection performance increased by 0.3\%. The mAP decreases by -2.7\%. 

\paragraph{AdaptiveShape anchor}
Introducing the AdaptiveShape grid generation improved the truck class by +10.9\% AP while all other classes are improved between 0.4\% and 0.9\%. 
This shows the importance of having multiple shape aware anchor distribution.

\paragraph{Standard ground truth augmentation}
Adding the standard ground truth augmentation for the truck class improves the truck metric by further +3.1\% AP. The other metrics decrease between -0.7\% and -1.2\% AP. One explanation for the decrease in performance for the non-truck class metrics is that the class distribution was further shifted towards large objects by the ground truth sampling within the backbone of the network. The mAP however increases by 0.1\%.

\paragraph{PairAugment}
Adding the pair database generation improves the truck metric by additional +2.2\% AP while the other metrics change between +0.2\% and -0.1\%. This shows how important it is to keep the pair relation for semi-trailer truck combinations. The mAP increases by 0.5\%.

\subsection{Fully correlated Gaussian heatmap}

Table \ref{tab:results_waymo} shows the results of differently sized fully correlated Gaussian heatmaps compared to the baseline from CenterFormer. We use two frames for temporal accumulation. 
The Gaussian scale in the table shows the length and width ratio used to scale the Gaussian standard deviations. 
The evaluation is done with our metric on the Waymo validation dataset subsampled using every 20th frame.
The benefits of having more positive anchors can be seen by the increase of mAP with larger Gaussian scale.

\begin{table}[t]
	\centering
	\begin{tabular}{lrr}
		\hline	
		Experiment    & Gaussian scale & mAP \\
		\hline
		CenterFormer  & -              & 78.0 \\
		AdaptiveShape & 1/12           & 78.1 \\
		AdaptiveShape & 1/8            & 79.2 \\
		AdaptiveShape & 1/6            & 79.3 \\
		\hline
	\end{tabular}
	\caption{
		Comparison of original CenterFormer implementation and ours on Waymo validation set with every 20. frame. Baseline with uncorrelated Gaussian distributions using the same standard deviations in x and y. Our implementation with fully correlated Gaussian distributions with standard deviations with differently scaled length and width ratios and appropriate rotation. 
	}
	\label{tab:results_waymo}
\end{table}

%-------------------------------------------------------------------------
\section{Conclusions}

We showed multiple strategies to improve the performance for 3D object detection
and shape estimation of arbitrary shaped vehicles. These new methods are especially beneficial for systems operating in logistic yards, harbors, airports, in mining or agricultural environments.
Adding the shape awareness to the underlying grid representations proofed to be most beneficial for the detection of large objects. Doing pairwise truck trailer augmentation further improved the performance.
Small objects did benefit most from the camera LiDAR fusion. However the improvement from camera fusion depends highly on the LiDAR resolution and number of object categories used.
Furthermore the choice of the metric plays an important role.

%-------------------------------------------------------------------------
{\small
	\bibliographystyle{ieee_fullname}
	\bibliography{egbib}

\begin{thebibliography}{10}\itemsep=-1pt

\bibitem{LongTailVisual}
Sherif Abdelkarim, Panos Achlioptas, Jiaji Huang, Boyang~Albert Li,
  Kenneth~Ward Church, and Mohamed Elhoseiny.
\newblock Long-tail visual relationship recognition with a visiolinguistic
  hubless loss.
\newblock {\em ArXiv}, abs/2004.00436, 2020.

\bibitem{nuScenes}
Holger Caesar, Varun Bankiti, Alex~H. Lang, Sourabh Vora, Venice~Erin Liong,
  Qiang Xu, Anush Krishnan, Yu Pan, Giancarlo Baldan, and Oscar Beijbom.
\newblock nuscenes: {A} multimodal dataset for autonomous driving.
\newblock {\em CoRR}, abs/1903.11027, 2019.

\bibitem{FeatureSpace}
Peng Chu, Xiao Bian, Shaopeng Liu, and Haibin Ling.
\newblock Feature space augmentation for long-tailed data.
\newblock {\em Lecture Notes in Computer Science}, page 694–710, 2020.

\bibitem{EffectiveSamples}
Yin Cui, Menglin Jia, Tsung-Yi Lin, Yang Song, and Serge~J. Belongie.
\newblock Class-balanced loss based on effective number of samples.
\newblock {\em 2019 IEEE/CVF Conference on Computer Vision and Pattern
  Recognition (CVPR)}, pages 9260--9269, 2019.

\bibitem{RangeDet}
Lue Fan, Xuan Xiong, Feng Wang, Nai long Wang, and Zhaoxiang Zhang.
\newblock Rangedet: In defense of range view for lidar-based 3d object
  detection.
\newblock {\em 2021 IEEE/CVF International Conference on Computer Vision
  (ICCV)}, pages 2898--2907, 2021.

\bibitem{AFDet}
Runzhou Ge, Zhuangzhuang Ding, Yihan Hu, Yu Wang, Sijia Chen, Li Huang, and
  Yuan Li.
\newblock Afdet: Anchor free one stage 3d object detection.
\newblock {\em ArXiv}, abs/2006.12671, 2020.

\bibitem{Kitti}
Andreas Geiger, Philip Lenz, Christoph Stiller, and Raquel Urtasun.
\newblock Vision meets robotics: The kitti dataset.
\newblock {\em The International Journal of Robotics Research}, 32:1231 --
  1237, 2013.

\bibitem{Sparse3dConv}
Benjamin Graham, Martin Engelcke, and Laurens van~der Maaten.
\newblock 3d semantic segmentation with submanifold sparse convolutional
  networks.
\newblock {\em 2018 IEEE/CVF Conference on Computer Vision and Pattern
  Recognition}, Jun 2018.

\bibitem{RareExamples}
Chiyu~Max Jiang, Mahyar Najibi, C. Qi, Yin Zhou, and Drago Anguelov.
\newblock Improving the intra-class long-tail in 3d detection via rare example
  mining.
\newblock {\em ArXiv}, abs/2210.08375, 2022.

\bibitem{PointPillars}
Alex~H. Lang, Sourabh Vora, Holger Caesar, Lubing Zhou, Jiong Yang, and Oscar
  Beijbom.
\newblock Pointpillars: Fast encoders for object detection from point clouds.
\newblock {\em CoRR}, abs/1812.05784, 2018.

\bibitem{CornerNet}
Hei Law and Jia Deng.
\newblock Cornernet: Detecting objects as paired keypoints.
\newblock {\em International Journal of Computer Vision}, 128(3):642–656, Aug
  2019.

\bibitem{Distillation}
Tianhao Li, Limin Wang, and Gangshan Wu.
\newblock Self supervision to distillation for long-tailed visual recognition.
\newblock {\em 2021 IEEE/CVF International Conference on Computer Vision
  (ICCV)}, pages 610--619, 2021.

\bibitem{FocalLoss}
Tsung{-}Yi Lin, Priya Goyal, Ross~B. Girshick, Kaiming He, and Piotr
  Doll{\'{a}}r.
\newblock Focal loss for dense object detection.
\newblock {\em CoRR}, abs/1708.02002, 2017.

\bibitem{BEVFusion}
Zhijian Liu, Haotian Tang, Alexander Amini, Xinyu Yang, Huizi Mao, Daniela Rus,
  and Song Han.
\newblock Bevfusion: Multi-task multi-sensor fusion with unified bird's-eye
  view representation.
\newblock {\em ArXiv}, abs/2205.13542, 2022.

\bibitem{AdamW}
Ilya Loshchilov and Frank Hutter.
\newblock Decoupled weight decay regularization, 2017.

\bibitem{HybridVoxel}
Jongyoun Noh, Sanghoon Lee, and Bumsub Ham.
\newblock Hvpr: Hybrid voxel-point representation for single-stage 3d object
  detection.
\newblock {\em 2021 IEEE/CVF Conference on Computer Vision and Pattern
  Recognition (CVPR)}, pages 14600--14609, 2021.

\bibitem{PointNetPlusPlus}
C. Qi, L. Yi, Hao Su, and Leonidas~J. Guibas.
\newblock Pointnet++: Deep hierarchical feature learning on point sets in a
  metric space.
\newblock In {\em NIPS}, 2017.

\bibitem{FrustumPointNet}
Charles~Ruizhongtai Qi, Wei Liu, Chenxia Wu, Hao Su, and Leonidas~J. Guibas.
\newblock Frustum pointnets for 3d object detection from {RGB-D} data.
\newblock {\em CoRR}, abs/1711.08488, 2017.

\bibitem{PointNet}
Charles~Ruizhongtai Qi, Hao Su, Kaichun Mo, and Leonidas~J. Guibas.
\newblock Pointnet: Deep learning on point sets for 3d classification and
  segmentation.
\newblock {\em CoRR}, abs/1612.00593, 2016.

\bibitem{PVRCNN}
Shaoshuai Shi, Chaoxu Guo, Li Jiang, Zhe Wang, Jianping Shi, Xiaogang Wang, and
  Hongsheng Li.
\newblock Pv-rcnn: Point-voxel feature set abstraction for 3d object detection.
\newblock {\em 2020 IEEE/CVF Conference on Computer Vision and Pattern
  Recognition (CVPR)}, pages 10526--10535, 2019.

\bibitem{PointRCNN}
Shaoshuai Shi, Xiaogang Wang, and Hongsheng Li.
\newblock Pointrcnn: 3d object proposal generation and detection from point
  cloud.
\newblock {\em 2019 IEEE/CVF Conference on Computer Vision and Pattern
  Recognition (CVPR)}, Jun 2019.

\bibitem{OneCycle}
Leslie~N. Smith and Nicholay Topin.
\newblock Super-convergence: very fast training of neural networks using large
  learning rates.
\newblock {\em Artificial Intelligence and Machine Learning for Multi-Domain
  Operations Applications}, May 2019.

\bibitem{Waymo}
Pei Sun, Henrik Kretzschmar, Xerxes Dotiwalla, Aurelien Chouard, Vijaysai
  Patnaik, Paul Tsui, James Guo, Yin Zhou, Yuning Chai, Benjamin Caine, Vijay
  Vasudevan, Wei Han, Jiquan Ngiam, Hang Zhao, Aleksei Timofeev, Scott
  Ettinger, Maxim Krivokon, Amy Gao, Aditya Joshi, Yu Zhang, Jonathon Shlens,
  Zhifeng Chen, and Dragomir Anguelov.
\newblock Scalability in perception for autonomous driving: Waymo open dataset.
\newblock {\em CoRR}, abs/1912.04838, 2019.

\bibitem{RangeSparseNet}
Pei Sun, Weiyue Wang, Yuning Chai, Gamaleldin~F. Elsayed, Alex Bewley, Xiao
  Zhang, Cristian Sminchisescu, and Drago Anguelov.
\newblock Rsn: Range sparse net for efficient, accurate lidar 3d object
  detection.
\newblock {\em 2021 IEEE/CVF Conference on Computer Vision and Pattern
  Recognition (CVPR)}, pages 5721--5730, 2021.

\bibitem{Attention}
Ashish Vaswani, Noam~M. Shazeer, Niki Parmar, Jakob Uszkoreit, Llion Jones,
  Aidan~N. Gomez, Lukasz Kaiser, and Illia Polosukhin.
\newblock Attention is all you need.
\newblock {\em ArXiv}, abs/1706.03762, 2017.

\bibitem{PointPainting}
Sourabh Vora, Alex~H. Lang, Bassam Helou, and Oscar Beijbom.
\newblock Pointpainting: Sequential fusion for 3d object detection.
\newblock {\em CoRR}, abs/1911.10150, 2019.

\bibitem{PointAugmenting}
Chunwei Wang, Chao Ma, Ming Zhu, and Xiaokang Yang.
\newblock Pointaugmenting: Cross-modal augmentation for 3d object detection.
\newblock In {\em Proceedings of the IEEE/CVF Conference on Computer Vision and
  Pattern Recognition (CVPR)}, pages 11794--11803, June 2021.

\bibitem{PillarBased}
Yue Wang, Alireza Fathi, Abhijit Kundu, David~A. Ross, Caroline Pantofaru,
  Thomas~A. Funkhouser, and Justin Solomon.
\newblock Pillar-based object detection for autonomous driving.
\newblock {\em CoRR}, abs/2007.10323, 2020.

\bibitem{LearningTail}
Yu-Xiong Wang, Deva Ramanan, and Martial Hebert.
\newblock Learning to model the tail.
\newblock In {\em NIPS}, 2017.

\bibitem{LearningExperts}
Liuyu Xiang and Guiguang Ding.
\newblock Learning from multiple experts: Self-paced knowledge distillation for
  long-tailed classification.
\newblock {\em ArXiv}, abs/2001.01536, 2020.

\bibitem{SecondNet}
Yan Yan, Yuxing Mao, and Bo Li.
\newblock Second: Sparsely embedded convolutional detection.
\newblock {\em Sensors}, 18(10), 2018.

\bibitem{HVNet}
Maosheng Ye, Shuangjie Xu, and Tongyi Cao.
\newblock Hvnet: Hybrid voxel network for lidar based 3d object detection.
\newblock {\em 2020 IEEE/CVF Conference on Computer Vision and Pattern
  Recognition (CVPR)}, pages 1628--1637, 2020.

\bibitem{CenterPoint}
Tianwei Yin, Xingyi Zhou, and Philipp Krahenbuhl.
\newblock Center-based 3d object detection and tracking.
\newblock {\em 2021 IEEE/CVF Conference on Computer Vision and Pattern
  Recognition (CVPR)}, Jun 2021.

\bibitem{CenterPointGithub}
Tianwei Yin, Xingyi Zhou, and Philipp Krahenbuhl.
\newblock Center-based 3d object detection and tracking.
\newblock \url{https://github.com/tianweiy/CenterPoint}, 2021.

\bibitem{Polarnet}
Yang Zhang, Zixiang Zhou, Philip David, Xiangyu Yue, Zerong Xi, and Hassan
  Foroosh.
\newblock Polarnet: An improved grid representation for online lidar point
  clouds semantic segmentation.
\newblock {\em 2020 IEEE/CVF Conference on Computer Vision and Pattern
  Recognition (CVPR)}, pages 9598--9607, 2020.

\bibitem{CenterNet}
Xingyi Zhou, Dequan Wang, and Philipp Krähenbühl.
\newblock Objects as points, 2019.

\bibitem{VoxelNet}
Yin Zhou and Oncel Tuzel.
\newblock Voxelnet: End-to-end learning for point cloud based 3d object
  detection.
\newblock {\em CoRR}, abs/1711.06396, 2017.

\bibitem{CenterFormer}
Zixiang Zhou, Xian Zhao, Yu Wang, Panqu Wang, and Hassan Foroosh.
\newblock Centerformer: Center-based transformer for 3d object detection.
\newblock In {\em European Conference on Computer Vision}, 2022.

\bibitem{CenterFormerGithub}
Zixiang Zhou, Xian Zhao, Yu Wang, Panqu Wang, and Hassan Foroosh.
\newblock Centerformer: Center-based transformer for 3d object detection.
\newblock \url{https://github.com/TuSimple/centerformer}, 2022.

\bibitem{ClassBalanced}
Benjin Zhu, Zhengkai Jiang, Xiangxin Zhou, Zeming Li, and Gang Yu.
\newblock Class-balanced grouping and sampling for point cloud 3d object
  detection.
\newblock {\em CoRR}, abs/1908.09492, 2019.

\bibitem{Cylindrical}
Xinge Zhu, Hui Zhou, Tai Wang, Fangzhou Hong, Yuexin Ma, Wei Li, Hongsheng Li,
  and Dahua Lin.
\newblock Cylindrical and asymmetrical 3d convolution networks for lidar
  segmentation.
\newblock {\em 2021 IEEE/CVF Conference on Computer Vision and Pattern
  Recognition (CVPR)}, pages 9934--9943, 2020.

\end{thebibliography}
}

\end{document}